\documentclass{article}
\usepackage{ijcai24}

\usepackage{times}
\usepackage{soul}
\usepackage{url}
\usepackage[hidelinks]{hyperref}
\usepackage[utf8]{inputenc}
\usepackage[small]{caption}
\usepackage{graphicx}
\usepackage{amsmath}
\usepackage{amsthm}
\usepackage{booktabs}
\usepackage{algorithm}
\usepackage{algorithmic}
\usepackage[switch]{lineno}
\usepackage{color,colortbl}
\usepackage{multirow}
\usepackage{subcaption}
\usepackage{tabularx}

\usepackage{subcaption}
\usepackage[export]{adjustbox}

\title{TFLOP: Table Structure Recognition Framework with Layout Pointer Mechanism}

\author{
Minsoo Khang
\And
Teakgyu Hong\\
\affiliations
Upstage AI, South Korea\\
\emails
\{mkhang, tghong\}@upstage.ai
}

\begin{document}

\maketitle

\begin{abstract}
    \textit{Table Structure Recognition (TSR)} is a task aimed at converting table images into a machine-readable format (e.g. HTML), to facilitate other applications such as information retrieval.
    Recent works tackle this problem by identifying the HTML tags and text regions, where the latter is used for text extraction from the table document. These works however, suffer from misalignment issues when mapping text into the identified text regions. In this paper, we introduce a new TSR framework, called TFLOP (\textbf{T}SR \textbf{F}ramework with \textbf{L}ay\textbf{O}ut \textbf{P}ointer mechanism), which reformulates the conventional text region prediction and matching into a direct text region pointing problem. Specifically, TFLOP utilizes text region information to identify both the table's structure tags and its aligned text regions, simultaneously. Without the need for region prediction and alignment, TFLOP circumvents the additional text region matching stage, which requires finely-calibrated post-processing. TFLOP also employs span-aware contrastive supervision to enhance the pointing mechanism in tables with complex structure. As a result, TFLOP achieves the state-of-the-art performance across multiple benchmarks such as PubTabNet, FinTabNet, and SynthTabNet. In our extensive experiments, TFLOP not only exhibits competitive performance but also shows promising results on industrial document TSR scenarios such as documents with watermarks or in non-English domain. Source code of our work is publicly available at: \url{https://github.com/UpstageAI/TFLOP}.
    
\end{abstract}

\section{Introduction}

\begin{figure}[t]
    \begin{subfigure}{.49\linewidth}
        \includegraphics[height=2.5in, left]{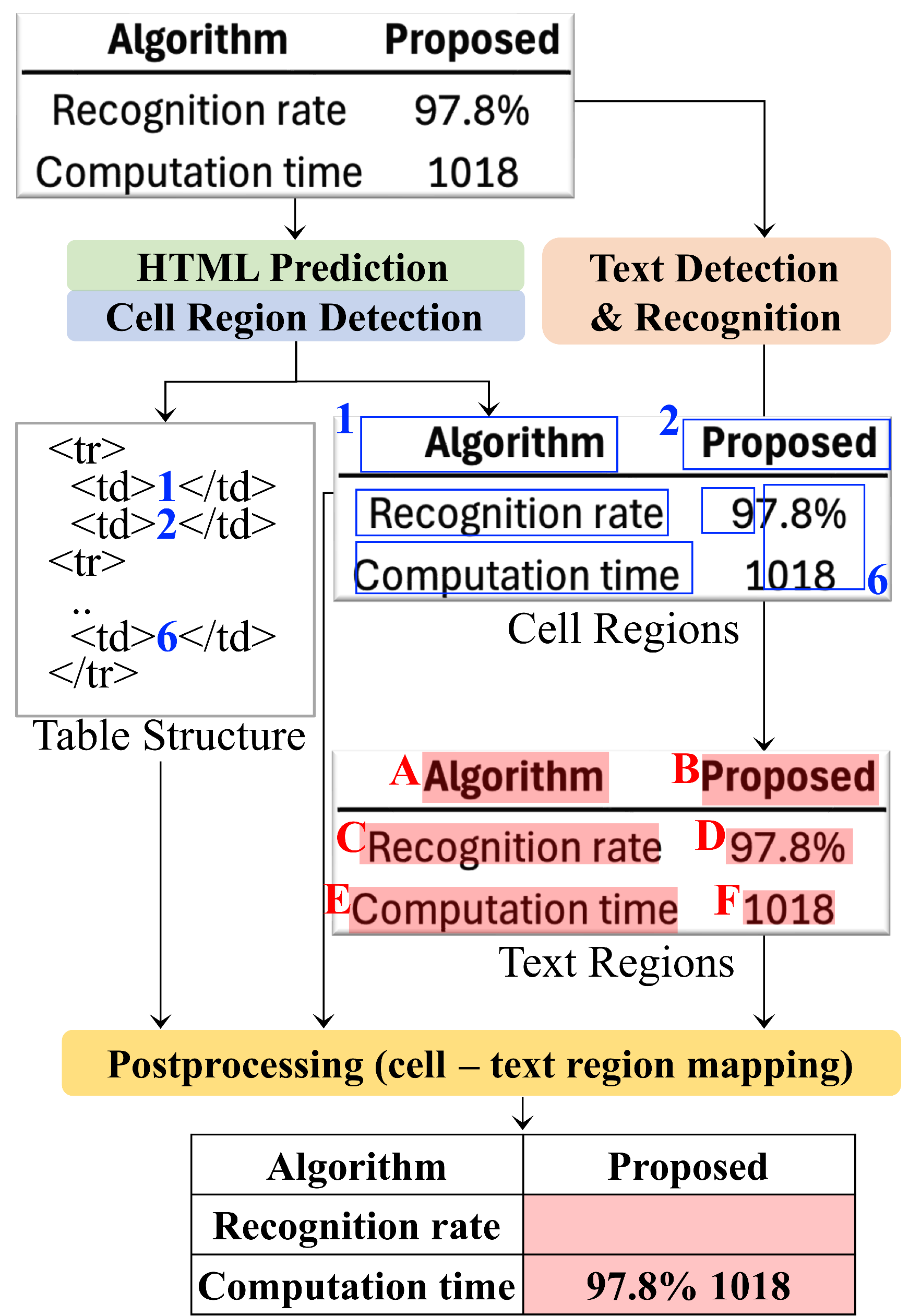}
        \caption{Dual decoder framework}
    \end{subfigure}
    \begin{subfigure}{.49\linewidth}
        \includegraphics[height=1.9in, right]{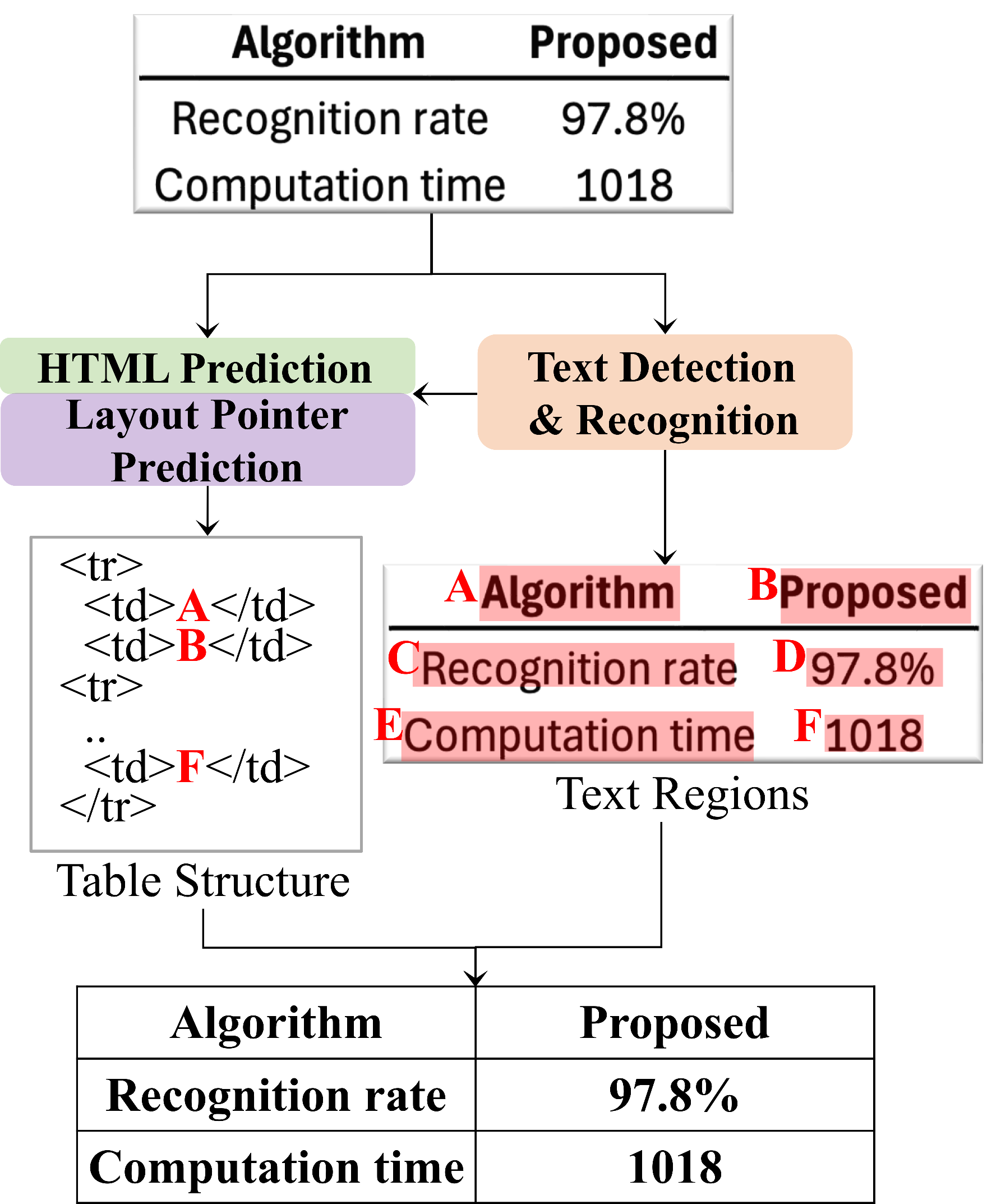}
        \caption{TFLOP}
    \end{subfigure}
    \caption{Overview of two TSR frameworks. The dual decoder identifies table cell regions and their HTML structure, requiring further cell and text region mapping for the final output. In contrast, TFLOP utilizes text region information and directly identifies the HTML structure with its corresponding text region relations.}
    \label{fig:main_idea}
\end{figure}

Tables are prevalent across a wide spectrum of documents (e.g. business documents, academic papers) for their compact and efficient representation. Such compact representation, however, presents a significant challenge for direct machine parsing. \textit{Table Structure Recognition} (TSR) aims to digitize table images into machine-readable format (e.g. HTML) representing their structure and text, allowing for various downstream applications such as information retrieval or table QA.

TSR often comprises of predicting two sets of structures before constructing the full table structure: logical and physical ~\cite{huang2023improving}. Logical structure represents the semantic organization and relational information between table cells, often represented in the form of HTML or LaTeX. Physical structure, on the other hand, represents the layout information of table cells such as their bounding boxes.

Recent works take on the image-to-text approach where both the logical and physical structures are predicted with the latter conditioning on the former. The physical structures (cell bounding boxes) are first mapped to text regions of the table, obtained using OCR engine or through PDF parsing, before combining the matched texts with the logical structure to form the full table. Despite its strong performance in logical structure prediction, these frameworks often suffer from misalignment issues where erroneous texts are matched due to imperfect alignment between the table text regions and the predicted cell bounding boxes. Such approaches require finely-calibrated post-processing during the matching of text regions for satisfactory results. 

This work, TFLOP, aims to eliminate the need for heuristic-based bounding box matching by leveraging table text regions directly in the framework through layout pointer mechanism. TFLOP reformulates the original bounding box prediction problem to a bounding box pointing problem. In particular, instead of predicting the cell bounding boxes conditioned on the logical structure, it predicts the associations between the bounding boxes and logical sequence through pointer mechanism. TFLOP's pointer mechanism not only serves as a remedy to the misalignment issue but also eliminates the need for heuristics-based bounding box matching.

On top of misalignment issues, recognizing structures of tables with row or column spans (i.e. complex tables) is one of the key challenges of TSR. Capitalizing on the flexibility of our framework, TFLOP employs span-aware contrastive supervision when processing the table text regions to improve its recognition of complex tables. Based on the proposed pointer mechanism and span-aware contrastive supervision, TFLOP achieves state-of-the-art performance across popular TSR benchmarks. 

In this work, we move beyond the benchmark datasets, and explore the versatility of TFLOP from the industrial perspective. We conduct extensive experiments and show that TFLOP not only has competitive performance but also the versatility in handling industrial document TSR scenarios such as watermarked documents or even non-English tables despite being trained exclusively on English tables.

The key contributions of our work are as follows:
\begin{itemize}
  \item We propose a novel TSR framework with layout pointer mechanism which not only remedies the text region misalignment issue but also eliminates the necessity for post-processing when mapping text regions into the predicted cell bounding boxes.
  
  \item We also present span-aware contrastive supervision in our framework. This supervision enhances the model's ability in recognizing structures of complex tables involving row or column spans.
  
  \item TFLOP achieves the state-of-the-art performance across multiple popular TSR benchmarks.

  \item Beyond the benchmark performance, TFLOP has also shown competitive performance and versatility when dealing with industrial TSR document scenarios such as tables with watermark or in non-English domain.
\end{itemize}

\section{Related Work}
TSR methods cover different variations of handling both logical and physical structure of tables. These methods can be largely categorised into two groups: detection-based and image-to-text methods.

\subsection{Detection-based TSR Methods}
Detection-based TSR is one of the common approaches that recognizes the table structure by leveraging on the table features detected such as separation lines or cell-level features. These methods typically proceed with physical structure understanding first before reasoning with the corresponding logical structure for TSR. 

\textbf{Grid-based approach} represents methods which utilise the grid representation based on the detected table features. Earlier works \cite{schreiber2017deepdesrt,paliwal2019tablenet} detects row and column masks through segmentation-based methods before aggregating them to form the table structure. SPLERGE~\cite{tensmeyer2019deep} then proposed split-and-merge pipeline which first detects the grid structure matching the table before merging adjacent cells to handle spanning entries. Follow-up works improved on top of this grid representation such as TRUST~\cite{guo2022trust} which proposed query-based splitting and vertex-based merging modules to improve spanning cell prediction, while SEM~\cite{zhang2022split} proposed aggregation of both visual and textual features in table grid generation. RobusTabNet~\cite{ma2023robust} proposed a spatial CNN module which improved physical structure reasoning when predicting separation lines prior to cell grid detection. Follow-up work TSRFormer~\cite{lin2022tsrformer} reformulated the line prediction task as a regression problem instead of image segmentation through a two-stage DETR~\cite{carion2020end} based approach. Recent work, GridFormer~\cite{lyu2023gridformer}, proposed a new method which directly predicts the vertexes and edges of the table grid (logical structure) from the table image.

\textbf{Cell-based approach} is another type of detection-based methods where cell-level features (physical structure) are first detected, before classifying the relation between cells (logical structure) to form the full table structure. Some of the representative works include TabStructNet~\cite{raja2020table} and FLAG-Net~\cite{liu2021show} which are end-to-end frameworks utilizing DGCNN architecture ~\cite{wang2019dynamic} to model the relation between the detected cell-level features. More recently, Hetero-TSR~\cite{liu2022neural} proposed NCGM which is designed to improve the cross-modality collaboration when handling complex TSR scenarios.

\begin{figure*}[t]
    \centering
    \includegraphics[width=0.9\linewidth]{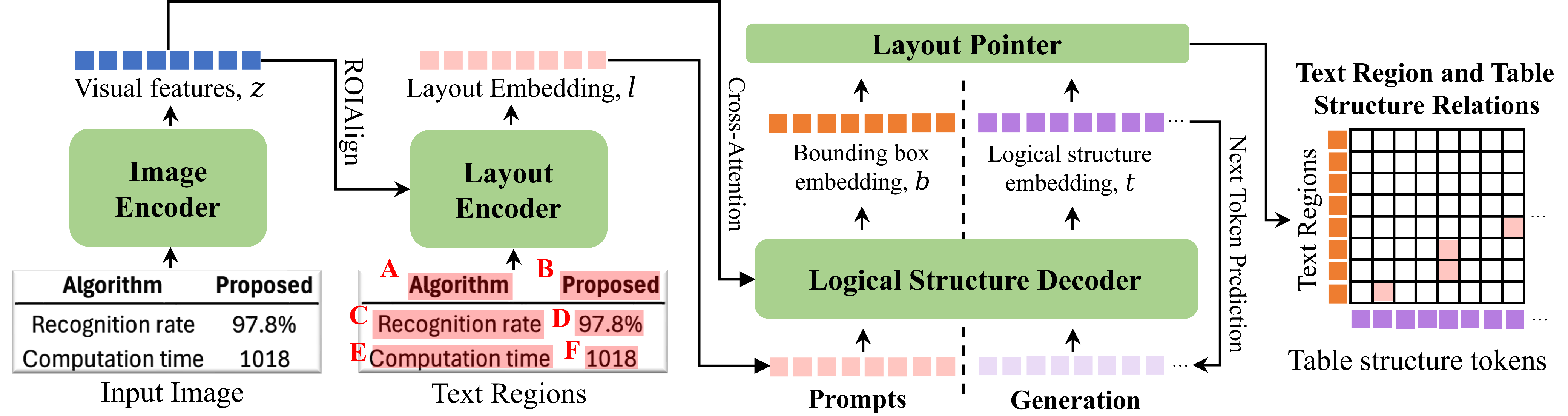}
    \caption{Overview illustration of TFLOP. Given a tabular image and its text region bounding boxes, visual features and layout embedding are output by the Image and Layout Encoders. Logical Structure Decoder then receives these features to auto-regressively generate table structure tokens (tags) while also predicting the associations between text region bounding boxes and table data tags through the Layout Pointer. These associations and table tags are aggregated to generate the full table structure.}
    \label{fig:model_pipeline}
\end{figure*}

\subsection{Image-to-Text based TSR Methods}
Image-to-Text methods reformulate the TSR task as an image-to-sequence translation task where the table structure is represented as a text sequence (e.g HTML, LaTeX, etc.). Recent methods typically predict the logical structure of the table first before conditioning on it for physical structure prediction. The two predictions are then aggregated to form the full table structure.

Earlier image-to-text TSR works directly produced the full table structure such as the work ~\cite{deng2019challenges} which modeled a LSTM-based table-to-LaTeX framework. Recent works on the other hand, transitioned to Transformer based sequence generation models which produced logical and physical structures separately before aggregating them for full table structure.

Notable examples of such works include ~\cite{ye2021pingan,nassar2022tableformer} which proposed Image Encoder Dual Decoder (IEDD) approach. In these works, after encoding the table image, HTML structure tags (logical structure) are first predicted by one of the decoders while the other conditioned on these tags to generate the cell bounding boxes (physical structure). These cell bounding boxes are subsequently mapped to text regions of tables obtained using OCR engines or through PDF parsing, before aggregating with the logical structure to complete the HTML sequence. 

Follow-up works proposed different means to improve the dual decoder framework. VAST~\cite{huang2023improving} proposed visual alignment loss to improve the physical structure prediction by enforcing detailed visual information in the decoding stage. Meanwhile, DRCC~\cite{shen2023divide} proposed a semi-autoregressive approach which reduced the effect of error accumulation in both logical and physical structure generation.

While both of these works' contributions do improve the structure predictions, they both suffer from the inherent issue of bounding box misalignment. When mapping the predicted cell bounding boxes for text retrieval, misalignment between the bounding boxes and text regions of tables could result in erroneous table structure. As such, physical structure prediction based frameworks are susceptible to bounding box misalignment issues and require heuristic post-processing for satisfactory results.

\section{Method}

\subsection{Overall Architecture}
TFLOP comprises of four modules: image encoder, layout encoder, logical structure decoder, and layout pointer. Our framework receives a table image and its corresponding text regions which are either provided in cell-level annotations or obtained using off-the-shelf OCR engines.

TFLOP first extracts the visual features from the table image using the image encoder while embedding the text region bounding boxes with the layout encoder. The generated visual features and layout embedding are then processed by the logical structure decoder. The visual features are provided to the decoder in a cross-attention mechanism while the layout embedding is processed as context prompt for generating the logical structure sequence.

On top of generating the logical structure (e.g. HTML-tags) auto-regressively, the decoder's last hidden state is further processed by the layout pointer module, which associates the predicted table data tags (e.g. \verb|<td>| for HTML, \textit{C} for OTSL) with the corresponding text regions to form the full table structure. TFLOP architecture is illustrated in Figure~\ref{fig:model_pipeline}.

\subsection{Image Encoder}
Motivated by the Donut architecture~\cite{kim2022ocr}, we use Swin Transformer ~\cite{liu2021swin} as TFLOP's image encoder.
All table images are preprocessed into a fixed resolution and embedded into visual features, $\{ z_i | z_i \in R^d, 1 \leq i \leq P \}$, where $P$ is the number of image patches and $d$ is the latent vector dimension.

\subsection{Layout Encoder}
Layout encoder comprises of MLP modules which embeds both the text region bounding boxes and the corresponding 2 x 2 ROIAlign~\cite{he2017mask} applied on the visual features, $\{z_i\}$. These embeddings are aggregated to form the layout embeddings, $\{ l_j | l_j \in R^{d}, 1 \leq j \leq B \}$, where $B$ is the context length for layout embedding.

\subsection{Logical Structure Decoder}
Logical structure decoder generates sequence of table tags conditioned on the visual features, $\{z_i\}$, and layout embedding, $\{l_j\}$.
TFLOP utilizes BART~\cite{lewis2019bart} architecture and follows configurations similar to that of Donut~\cite{kim2022ocr}.
TFLOP's decoder outputs a sequence of $\{y_k | y_k \in R^v, \ 1 \leq k \leq T\}$ where $T$ is the total number of table tags and $v$ is the token vocabulary size. Cross-entropy loss, $\mathcal{L}_{cls}$, is employed to supervise the decoder's tag classification.

Prior works' decoders ~\cite{shen2023divide,huang2023improving,nassar2022tableformer} generate logical structure sequence in the HTML format. Despite its long sequence, HTML representation is often used for its flexibility and wide coverage of tabular layouts. To reduce its long sequence length, ~\cite{huang2023improving,ye2021pingan} merged specific tags (e.g. \verb|<td></td>|). TFLOP achieves similar effect by generating OTSL-tag sequences~\cite{lysak2023optimized} which have 1-to-1 mapping with the target HTML sequence.

\subsection{Layout Pointer}
Apart from generating a sequence of table tags, the decoder's last hidden state features, $\{h_i | h_i \in R^d, \ 1 \leq i \leq N\}$, are used in our layout pointer module. $N$ is the sum of the number of bounding boxes ($B$) and the number of table tags ($T$).
Specifically, the feature sequence, $\{h_i\}_{i=1}^{N}$, is first split into two sub-sequences: $\{b_j\}_{j=1}^{B}$ and $\{t_k\}_{k=1}^{T}$.
$\{b_j\}_{j=1}^{B}$ is a sequence of fixed-length $B$, representing the last hidden state features of the bounding boxes.
$\{t_k\}_{k=1}^{T}$, on the other hand, is a sequence of length $T$, representing the last hidden state features of the predicted table tags.
These two sequence of features are then projected into $\{ \bar{b}_j \}$ and $ \{ \bar{t}_k \}$ through linear transformation (Equation~\ref{eqn:bbox_proj}).
Among the table tag features, $\{ \bar{t}_k \}$, we define the indices of those which correspond to table data tags as set $D$.
Layout pointer supervision is then applied as in Equation~\ref{eqn:pointerloss}.
\begin{equation}
    \label{eqn:bbox_proj}
    \bar{b}_j = \text{proj}_{b}(b_j),
    \quad
    \bar{t}_k = \text{proj}_{t}(t_k)
\end{equation}
\begin{equation}
    \label{eqn:pointerloss}
    \mathcal{L}_{ptr}=-\frac{1}{B}\sum_{j=1}^{B}\log(\frac{\exp(\bar{b}_j \cdot \bar{t}_{k^*} / \tau)}{\sum_{k^\prime\in D}\exp(\bar{b}_j \cdot \bar{t}_{k^\prime} / \tau)})
\end{equation}

$\mathcal{L}_{ptr}$ represents the loss for layout pointer supervision where $\bar{b}_j$ represents the projected feature of the $j^{th}$ bounding box and $k^*$ is the index of the table data tag corresponding to the $j^{th}$ bounding box. $\cdot$ and $\tau$ denote the dot product and the temperature hyper-parameter, respectively. It is worth noting that, the bounding box and the table tags have a one-to-one or a many-to-one relation as there could be one or more text bounding boxes present within a single table cell. As such, in Equation~\ref{eqn:pointerloss}, $\mathcal{L}_{ptr}$ is calculated by evaluating the negative log-likelihood for each of the $B$ bounding boxes before taking their arithmetic mean.

It should also be noted that, it is possible for table data tags to not have any corresponding bounding boxes (i.e. empty table cell). To ensure provision of pointer supervision for all table data tags, a separate loss supervision $\mathcal{L}_{ptr}^{empty}$ is applied to those without any corresponding bounding boxes as following:
\begin{equation}
    \label{eqn:emptypointerloss}
    \mathcal{L}_{ptr}^{empty}=-\frac{1}{|D|}\sum\limits_{k^\prime\in D}\text{BCE}(\sigma(\bar{b}_0\cdot \bar{t}_{k^\prime}), \text{I}(k^\prime))
\end{equation}

$\bar{b}_0$ is the linear projection of a special embedding dedicated to empty table data tags. $\sigma()$ and $\text{BCE}()$ represents sigmoid activation function and Binary Cross-Entropy, respectively, while $\text{I}(k^\prime)$ represents binary label indicating whether $k^\prime$ data tag is empty.

\subsection{Span-aware Contrastive Supervision}
To better address complex table structures (with rowspan or colspan), TFLOP adopts span-aware contrastive supervision across the bounding box embeddings, $\{ b_j \}$, to improve its tabular layout understanding.
While prior works provide contrastive supervision on table elements both row-wise and column-wise, TFLOP takes a step further by introducing span-aware adjustments to this supervision.

\begin{figure}[t]
    \centering
    \includegraphics[width=\linewidth]{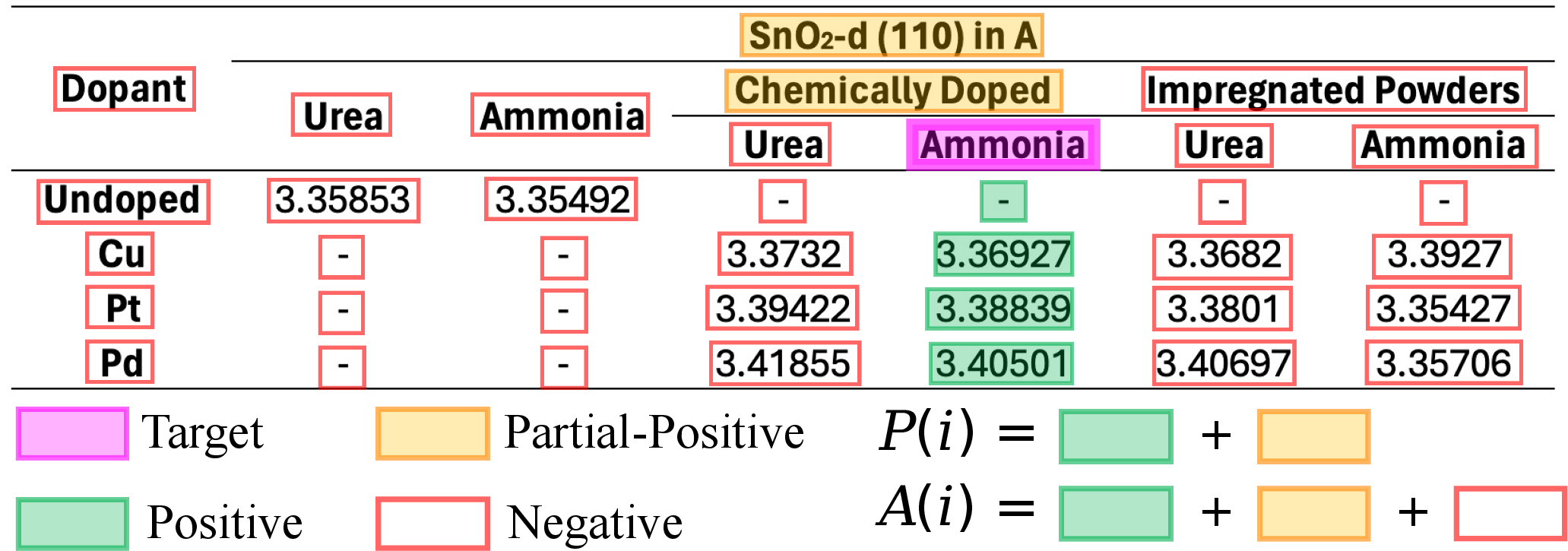}
    \caption{Sample visualisation of span-aware constrastive supervision involving multi-span structures. In the column-wise contrastive supervision example above, for a given bounding box ($i$, pink), positive samples ($P(i)$) are those with either full overlap (green) or partial overlap (orange), while the rest (red) are negative samples.}
    \label{fig:spanaware_idea}
\end{figure}

Given a $j^{th}$ bounding box embedding $b_j$, it is first projected using a linear layer to form $\hat{b}_j$ (Equation~\ref{eqn:contrastive_projection}), before evaluating its span-aware contrastive loss as shown in Equation~\ref{eqn:spanaware_contrastive}.
\begin{equation}
    \label{eqn:contrastive_projection}
    \hat{b}_j=\text{proj}_{s}(b_j)
\end{equation}
\begin{equation}
    \label{eqn:spanaware_contrastive}
    \resizebox{0.91\linewidth}{!}{$
        \mathcal{L}_{contr,j} = -\frac{1}{\sum\limits_{p\in P(j)}c_p(j)} \sum_{p\in P(j)}c_p(j)\log(\frac{\exp(\hat{b}_j \cdot \hat{b}_p / \tau)}{\sum\limits_{a\in A(j)}\exp(\hat{b}_j \cdot \hat{b}_a / \tau)})
    $}
\end{equation}

$\mathcal{L}_{contr,j}$ represents the span-aware contrastive loss for the $j^{th}$ bounding box. This formulation is applicable to both row-span and column-span supervision. Here, $A(j)$ represents all the bounding boxes except the $j^{th}$, $P(j)$ represents all the bounding boxes of $A(j)$ that are positive samples (i.e. either same row or column as the $j^{th}$ bounding box), and $\hat{b}_p$ and $\hat{b}_a$ represent the projected bounding box embedding of $P(j)$ and $A(j)$, respectively.
The above formulation follows similar to that of Supervised Contrastive Loss ~\cite{khosla2020supervised} except for the span-coefficient $c_p(j)$.

Span-coefficient $c_p(j)$ denotes the degree of proximity between $j^{th}$ and $p$ based on the span overlap between the two bounding boxes. For example, with reference to Figure~\ref{fig:spanaware_idea}, in column-wise contrastive supervision, the span-coefficient between the $j^{th}$ bounding box (pink) and a positive bounding box (green or yellow) can be formulated as:
\begin{equation}
    \label{eqn:span-coefficient}
    c_p(j)=\frac{(\text{overlap}(p, j))^2}{\text{span}(p) \times \text{span}(j)}
\end{equation}

Here, $\text{span}()$ denotes the span count (either row or column) for the given bounding box, while $\text{overlap}(x,y)$ denotes the number of overlap cells between the bounding box $x$ and $y$. For example, span-coefficient of the bounding box of ``Ammonia'' against that of ``Chemically Doped'' in Figure~\ref{fig:spanaware_idea} would be $1/(2\times1)$.

It is worth noting that, when the span-coefficient is set to a constant value of $1$ (i.e. uniform contrastive supervision), $\mathcal{L}_{contr,j}$ reduces to the standard supervised contrastive loss formulation ~\cite{khosla2020supervised}.

\subsection{Loss Function}
TFLOP's training objective is composed of tag classification loss, layout pointer loss, and span-aware contrastive loss.
Tag classification loss ($\mathcal{L}_{cls}$) is evaluated using the negative-log likelihood of the table tag predictions, while layout pointer loss is a linear combination of $\mathcal{L}_{ptr}$ and $\mathcal{L}_{ptr}^{empty}$.
Span-aware contrastive loss is also a linear combination of $\mathcal{L}_{contr,j}^{row}$ and $\mathcal{L}_{contr,j}^{col}$ which denote row-wise and column-wise contrastive loss for the $j^{th}$ bounding box respectively.
\begin{align}
\begin{aligned}
    \mathcal{L} & = \lambda_{1}\mathcal{L}_{cls} +
    \lambda_{2}\mathcal{L}_{ptr} + \lambda_{3}\mathcal{L}_{ptr}^{empty} \\
    & + \lambda_{4}\frac{1}{B}\sum_{j=1}^B \mathcal{L}_{contr,j}^{row} + \lambda_{5}\frac{1}{B}\sum_{j=1}^B \mathcal{L}_{contr,j}^{col} 
\end{aligned}\label{eqn:loss_formulation}
\end{align}

\begin{table}[t]
    \tabcolsep = 3pt
    \centering
    \begin{tabular}{lccccc}
    \toprule
        \multirow{2}{*}{Methods} & \multicolumn{2}{c}{\textbf{PubTabNet}.Val} && \multicolumn{2}{c}{\textbf{PubTabNet}.Test} \\
        \cmidrule{2-3}\cmidrule{5-6}
        & TEDS-S & TEDS && TEDS-S & TEDS\\
        \midrule
        TableMaster~\shortcite{ye2021pingan} & - & - && - & 96.32 \\
        LGPMA~\shortcite{qiao2021lgpma} & 96.7 & 94.6 && - & - \\
        TableFormer~\shortcite{nassar2022tableformer} & 97.5 & - && 96.75 & 93.60\\
        VAST~\shortcite{huang2023improving} & - & - && 97.23 & 96.31\\
        RobusTabNet~\shortcite{ma2023robust} & 97.0 & - && - & -  \\
        DRCC~\shortcite{shen2023divide} & \textbf{98.9} & \underline{97.8} && - & - \\
        \midrule
        TFLOP$_{\text{BASE}}$ & 98.1 & \underline{97.8} && \underline{98.25} & \underline{96.42}\\
        TFLOP$_{\text{FULL}}$ & \underline{98.3} & \textbf{98.0} && \textbf{98.38} & \textbf{96.66}\\
        \bottomrule
    \end{tabular}
    \caption{TEDS-Struct (TEDS-S) and TEDS evaluation on PubTabNet validation and test dataset. 
    }
    \label{tbl:pubtabnet_all}
\end{table}

\section{Experiments}

\subsection{Datasets}
To validate the effectiveness of our framework, experiments are conducted against three popular TSR benchmark datasets: PubTabNet~\cite{zhong2020image}, FinTabNet~\cite{zheng2021global}, and SynthTabNet~\cite{nassar2022tableformer}.

\textbf{PubTabNet} is one of the large-scale TSR datasets containing HTML annotations of tables extracted from scientific articles. It is composed of 500,777 training and 9,115 validation table images. Annotated test dataset comprising of 9,064 images was subsequently released, and TFLOP's TSR performance against both the validation and test datasets are reported in this work. It should be noted that for PubTabNet test dataset, no cell-level annotation (i.e. text region bounding box) is provided and off-the-shelf OCR engine was used to obtain these annotations.

\textbf{FinTabNet} is one of the popular TSR benchmarks composed of single-page PDF documents from financial reports. This dataset comprises of 112,887 tables extracted from the documents along with the cell-level annotations. FinTabNet facilitates the evaluation of TFLOP's performance in tables where the text regions are not obtained using OCR engines (i.e. free from OCR-related noise similar to PDF parsing).

\textbf{SynthTabNet} was introduced by ~\cite{nassar2022tableformer} as a benchmark dataset that is not only large-scale but also diverse in table appearances and content. SynthTabNet is composed of 600,000 table images across different styles and provides cell-level annotation similar to that of FinTabNet.

\subsection{Experimental Settings}
In training of TFLOP, input image resolution is set to $768\times768$ across all benchmark datasets. The output sequence length, $N$, is fixed at 1,376 to allow sufficient length for the layout embedding and generation of the table tags. Feature dimension $d$ of the framework is set to 1,024 and the hyper-parameters of the loss formulation Equation~\ref{eqn:loss_formulation} are: $\lambda_{1}=\lambda_{2}=\lambda_{3}=1$ and $\lambda_{4}=\lambda_{5}=0.5$. The temperature value $\tau$ is set to 0.1. All experiments were conducted with 4$\times$A100 GPUs at 250K training steps.

\subsection{Evaluation Metrics}
To evaluate TFLOP's performance, we utilize Tree-Edit-Distance-Based Similarity, TEDS~\cite{zhong2020image}, and TEDS-Struct~\cite{huang2023improving,nassar2022tableformer} which computes the TEDS score between the predicted and ground truth HTML table structure with and without table text content, respectively.
\begin{equation}
    \label{eqn:TEDS}
    \text{TEDS}(T_{pr},T_{gt})=1-\frac{\text{EditDist}(T_{pr},T_{gt})}{\max(\lvert T_{pr}\rvert,\lvert T_{gt}\rvert)}
\end{equation}

In Equation~\ref{eqn:TEDS}, $T$ and $\lvert T\rvert$ represent the HTML structure and number of nodes in $T$, respectively, while $\text{EditDist}()$ indicates tree-edit distance between the HTML structures.

\subsection{Results}
We benchmarked TFLOP against three popular datasets as shown in Tables~\ref{tbl:pubtabnet_all} and~\ref{tbl:fin_synth_table}. For all benchmarks, we not only report the results of TFLOP$_{\text{FULL}}$, but also TFLOP$_{\text{BASE}}$ to better evaluate the effectiveness of our layout pointer mechanism. TFLOP$_{\text{BASE}}$ differs from TFLOP$_{\text{FULL}}$ with the absence of image ROIAlign and span-aware contrastive supervision.

\begin{table}[t]
    \tabcolsep = 3pt
    \centering
    \begin{tabular}[h]{lccccc}
        \toprule
        \multirow{2}{*}{Methods} & \multicolumn{2}{c}{\textbf{FinTabNet}} && \multicolumn{2}{c}{\textbf{SynthTabNet}} \\
        \cmidrule{2-3} \cmidrule{5-6}
        & TEDS-S & TEDS && TEDS-S & TEDS \\
        \midrule
        TableFormer~\shortcite{nassar2022tableformer} & 96.80 & - && 96.70 & - \\
        GridFormer~\shortcite{lyu2023gridformer} & 98.63 & - && - & - \\
        VAST~\shortcite{huang2023improving} & 98.63 & 98.21 && - & - \\
        DRCC~\shortcite{shen2023divide} & - & - && 98.70 & -   \\
        \midrule
        TFLOP$_{\text{BASE}}$ & \underline{99.43} & \underline{99.22} && \textbf{99.42} & \underline{99.34}\\
        TFLOP$_{\text{FULL}}$ & \textbf{99.56} & \textbf{99.45} && \textbf{99.42} & \textbf{99.40} \\
        \bottomrule
    \end{tabular}
    \caption{TEDS-Struct / TEDS on FinTabNet and SynthTabNet.}
\label{tbl:fin_synth_table}
\end{table}

\begin{figure}[t]
    \centering
    \includegraphics[width=\linewidth]{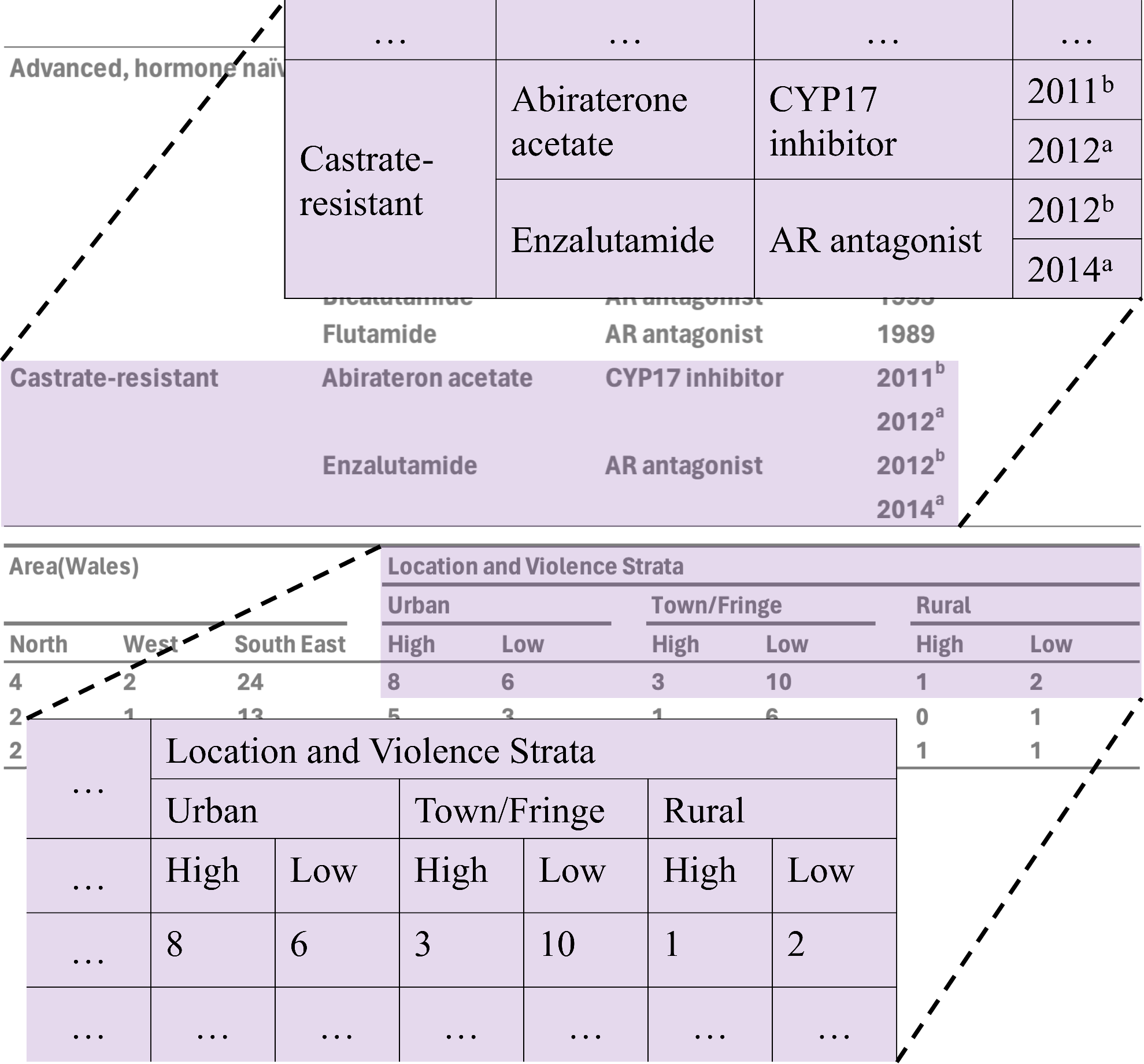}
    \caption{Visualisations of tables constructed from generated HTML sequences with corresponding tabular images (recreated for improved legibility) for reference. TFLOP successfully constructs tables with complex structures such as hierarchical row-spans (top) or hierarchical column-spans (bottom).}
    \label{fig:pubtabnet_qualitative}
\end{figure}

Table~\ref{tbl:pubtabnet_all} results show that TFLOP outperforms prior works in recognition of full table structure across the validation and test splits of PubTabNet. Evaluation across  PubTabNet's validation dataset was conducted to ensure fair comparisons against prior works which only reported results on the validation set. For the test dataset, where cell-level annotations are not provided, text region bounding box annotations were obtained using PSENet~\cite{wang2019shape} and Master~\cite{lu2021master} similar to prior works ~\cite{ye2021pingan,guo2022trust,huang2023improving} for fair comparisons. TFLOP's state-of-the-art performance on PubTabNet's test dataset clearly demonstrates our framework's efficacy when using text regions derived from off-the-shelf OCR engines. Visualisations in Figure~\ref{fig:pubtabnet_qualitative} illustrate TFLOP's ability to recognize tables with complex structures such as hierarchical row-spans (top) and hierarchical column-spans (bottom).

Results in Table~\ref{tbl:fin_synth_table} also substantiate TFLOP's superior performance over various prior works by achieving state-of-the-art recognition results for both FinTabNet and SynthTabNet. FinTabNet being table extracts from financial reports, TFLOP's state-of-the-art performance (\textbf{99.45} TEDS) has significant implications in the context of industrial applications where the margin for error is exceedingly narrow. SynthTabNet, on the other hand, comprises of table structures of various styles and TFLOP's state-of-the-art performance (\textbf{99.40} TEDS) clearly demonstrates that the framework is not restrictive to specific tabular format or style.

In both Tables~\ref{tbl:pubtabnet_all} and~\ref{tbl:fin_synth_table}, it can be noted that TFLOP also achieves significant improvement in terms of TEDS-Struct metric (HTML table tags only). We posit that this is a side-effect of layout embedding in our framework. Provision of layout embedding is essential for our framework's layout pointer mechanism as it serves as the pointing target from the generated table tags. Incidentally, this layout embedding could also improve the framework's understanding of the table's layout, resulting in improved table tag generation as shown by TFLOP's TEDS-Struct results.

\begin{table}[t]
    \tabcolsep = 3pt
    \centering
    \begin{tabular*}{0.9\hsize}{@{\extracolsep{\fill}}lcccc}
    \toprule
        Methods & Simple & Complex & All \\
        \midrule
        TFLOP$_{\text{BASE}}$ & 97.92 & 94.85 & 96.42\\
        TFLOP$_{\text{BASE}}$ + I & +0.04 & +0.14 & +0.08\\
        TFLOP$_{\text{BASE}}$ + I + U & +0.01 & +0.12 & +0.06\\
        TFLOP$_{\text{BASE}}$ + I + S & \textbf{+0.14} & \textbf{+0.35} & \textbf{+0.24}\\
        \midrule\
        
        TFLOP$_{\text{FULL}}$ & \textbf{98.06} & \textbf{95.20} & \textbf{96.66}\\
        \bottomrule
    \end{tabular*}
    \caption{Ablation of I(ImageROI), U(Uniform contrastive) and S(Span-aware contrastive) on PubTabNet test Dataset in TEDS (\%). Note that TFLOP$_{\text{BASE}}$ + I + S and TFLOP$_{\text{FULL}}$ are equivalent.
    }
    \label{tbl:pubtabnet_ablation}
\end{table}

On top of achieving the state-of-the-art TEDS score across benchmark datasets, it is also worth noting of the gap between TEDS-Struct and TEDS scores of our framework in comparison to prior works. While the TEDS metric evaluates the accuracy of the full table structure, the gap between TEDS-Struct and TEDS serves as an indirect indication for significance of bounding box misalignments (for prior works) or significance of layout pointer mechanism (for TFLOP). Aside from PubTabNet test dataset where the TEDS metric is also affected by the OCR error of PSENet~\cite{wang2019shape} and Master~\cite{lu2021master}, TFLOP consistently achieves the smallest gap between TEDS-Struct and TEDS across remaining benchmarks (e.g. \textbf{0.11} vs 0.42 in FinTabNet).
This clearly shows the effectiveness of layout pointer mechanism in addressing bounding box misalignments faced by prior works.

\begin{figure}[t]
    \centering
    \includegraphics[width=\linewidth]{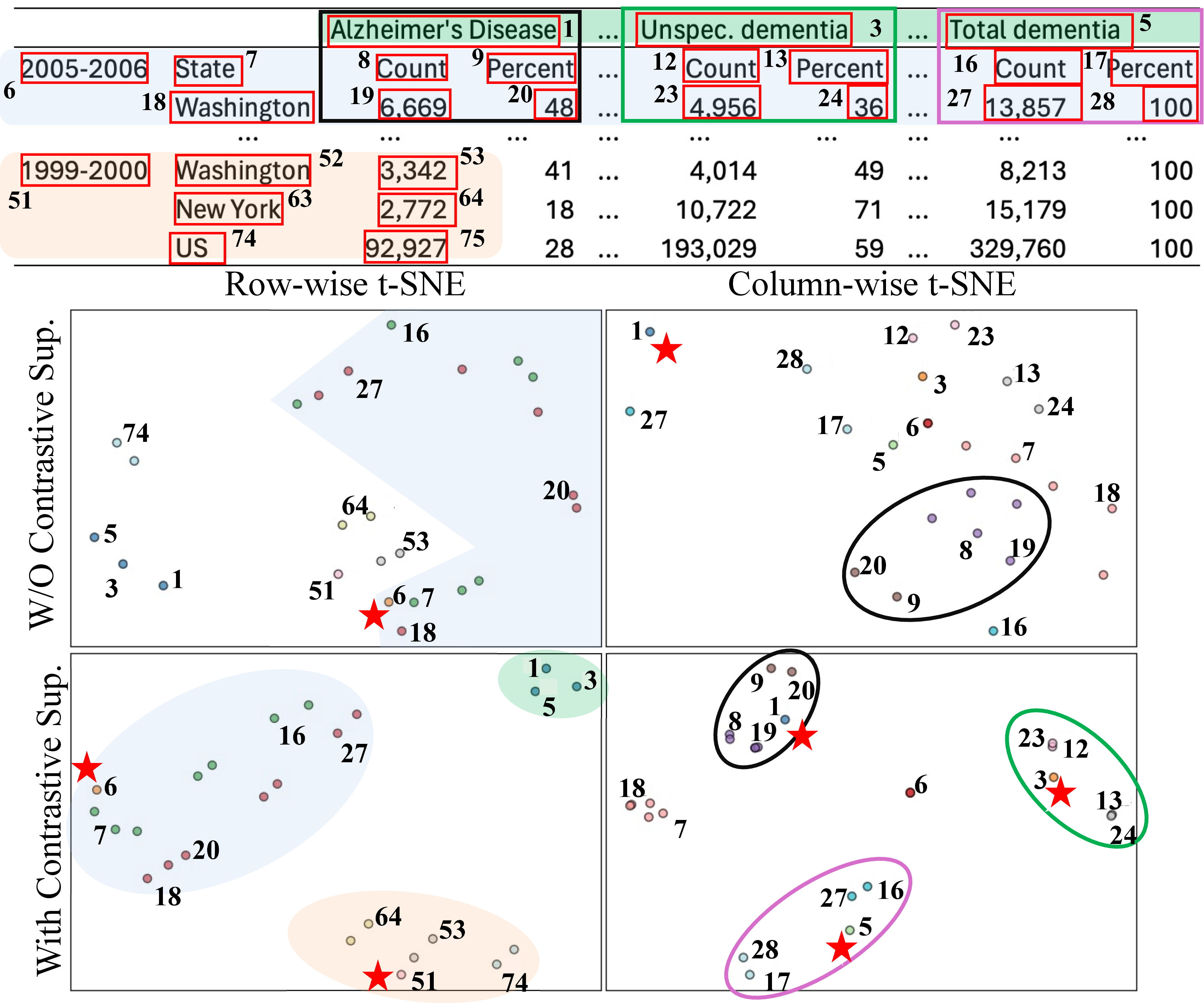}
    \caption{Row-wise and column-wise t-SNE visualisation of bounding box embeddings. PubTabNet table image (top, table recreated for improved legibility) and 25 bounding boxes are sampled for visualisation. Filled-colours represent different row-span groups while border-colours represent different column-span groups sampled for visualisation. Colours in t-SNE plots match that of table above and boxes spanning multi-rows/columns are marked with a red star.}
    \label{fig:vector_image}
\end{figure}

\subsection{Ablation Study}

Aside from layout pointer mechanism, we analyzed the effectiveness of other components in our framework by comparing between TFLOP$_{\text{BASE}}$ and TFLOP$_{\text{FULL}}$ in Table~\ref{tbl:pubtabnet_ablation}. Firstly, for image ROI alignment, consistent with ~\cite{huang2023improving,shen2023divide}, it is evident from Table~\ref{tbl:pubtabnet_ablation} that incorporating RoI aligned visual features into layout embedding is also beneficial for recognizing table structures in our framework. Secondly, Table~\ref{tbl:pubtabnet_ablation} shows clear performance improvement through span-aware contrastive supervision over other method configurations. The performance gain is most noticeable for tables with complex structure, showing that our span-aware contrastive supervision benefits the framework with improved recognition of tables with row or column spans. This can also be observed in t-SNE visualisation of bounding box embedding space (Figure~\ref{fig:vector_image}) where, embeddings are distinctly structured into clusters of row or column spans.

\section{TFLOP Versatility}
On top of its strong TSR performance, we further explore the versatility of our framework in two scenarios commonly encountered during industrial application of TSR: tables with watermark and non-English texts.

\subsection{Watermark TSR}

Unlike the benchmark dataset tables, tables in real industrial documents often contain unwanted texts such as watermark. These unwanted texts could result in recognition of erroneous table structure if not filtered accurately.
\begin{figure}[t]
    \centering
    \includegraphics[width=\linewidth]{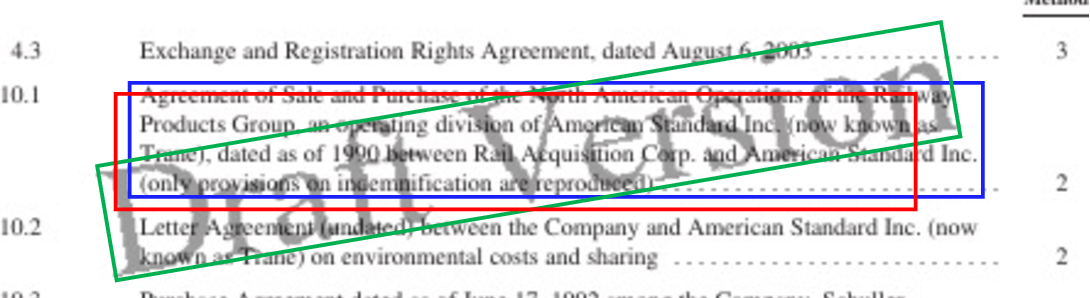}
    \caption{Sample FinTabNet image with a ``Draft Version" watermark demonstrates the challenge of processing watermarks in TSR using a dual-decoder framework. Blue and green boxes indicate text regions and watermark areas, while red shows a sample prediction.}
    \label{fig:watermark_idea_figure}
\end{figure}

Prior works based on the dual-decoder framework are not optimal in handling tables with watermark as they require complex bounding box matching heuristics to properly distinguish wanted text region bounding boxes from those of watermark (as illustrated in Figure~\ref{fig:watermark_idea_figure}). Our work, TFLOP, on the contrary, has the versatility to be trained to ignore these watermark bounding boxes prior to layout pointing.

To support this, we first prepared watermark table dataset by inpainting watermarks into the FinTabNet~\cite{zheng2021global} dataset. We then trained TFLOP with this dataset, requiring only a minor addition of a two-layer MLP with a binary cross-entropy loss function. In brief, prior to predicting pointer associations between bounding boxes and table tags, a binary classifier is trained to filter watermark bounding boxes. More details can be found in supplementary material.

In Table~\ref{tbl:watermark}, we compare TFLOP's TSR on watermark dataset against TableMaster and variations of gold annotation which assume error-free logical structure. Gold$_{\text{greedy}}$ constructs full table structure by including all watermarks that has any bounding box IOU with the text bounding boxes, while Gold$_{\text{selective}}$ filters watermarks with IOU threshold of 0.5. Table~\ref{tbl:watermark} shows promising result where TFLOP filters out most of the watermark texts with just a simple addition of two-layer MLP, showcasing the versatility of our framework.

\begin{table}[t]
    \tabcolsep = 3pt
    \centering
    \begin{tabular*}{0.9\hsize}{@{\extracolsep{\fill}}lcccc}
    \toprule
        Methods & IOU & TEDS-Struct (\%) & TEDS (\%) \\
        \midrule
        TableMaster & - & 82.18 & 72.83\\
        Gold$_{\text{greedy}}$ & 0.0 & - & 96.45 \\
        Gold$_{\text{selective}}$ & 0.5 & - & 98.16 \\
        \midrule\
        TFLOP$_{\text{FULL}}$ & - & \textbf{99.54} & \textbf{99.41}\\
        \bottomrule
    \end{tabular*}
    \caption{TSR performance on watermarked FinTabNet dataset.
    }
    \label{tbl:watermark}
\end{table}

\subsection{Cross-lingual TSR}
Another important aspect for industrial TSR applications lies in the performance across non-English tables. TSR on non-English tables is a challenging task due to limited availability of data and thus, we examine the versatility of TFLOP in performing TSR on non-Engligh tables despite having only been trained on English tables. For this purpose, we self-annotated 30 Korean table images (15 simple \& 15 complex tables) extracted from real Korean financial reports, including both the HTML sequence and cell-level annotations of tabular images.

We benchmarked our framework against TableMaster~\cite{ye2021pingan} and GPT-4V~\cite{openai2023gpt-4v}. It should be noted that both TFLOP and TableMaster were trained on PubTabNet dataset prior to evaluating on the Korean table dataset. Results in Table~\ref{tbl:KoreanTable} show promising results demonstrating the cross-lingual versatility of TFLOP by outperforming both TableMaster and GPT-4V consistently across simple and complex Korean tables by a significant margin.

To better under the industrial implications of TSR results in Table~\ref{tbl:KoreanTable}, we conducted an additional \textit{Question-Answering} (QA) assessment on top of the generated table structures. For the assessment, we built 175 unique question-answer pairs where the questions require clear understanding of the table provided to answer accurately. In the assessment, both the question and table structure generated (HTML) are provided to GPT-4V along with the tabular image, before comparing its output with the answer label. Each of the 175 answers were evaluated manually for the QA accuracy shown in Table~\ref{tbl:KoreanTable}.

QA accuracy results in Table~\ref{tbl:KoreanTable} not only show the importance of HTML sequence for Table QA in non-English domain for GPT-4V, but also highlight how the difference in TEDS score could translate into the real industrial application of table QA in cross-lingual setting. Details of the dataset and qualitative results can be found in supplementary material.

\begin{table}[t]
\tabcolsep = 3pt
\centering
\begin{tabular*}{0.9\hsize}{@{\extracolsep{\fill}}lccc}
\toprule
\multirow{2}{*}{Methods} & \multicolumn{2}{c}{TEDS (\%)} & \multirow{2}{*}{QA Acc. (\%)} \\
                         & Simple        & Complex       &                               \\
\midrule
Image-only               & -             & -             & 56.00                         \\
GPT-4V                   & 79.43         & 68.39         & 78.86                         \\
TableMaster              & 89.96         & 83.94         & 82.86                         \\
\midrule
TFLOP                    & \textbf{95.76}         & \textbf{89.41}         & \textbf{92.00}                        \\
\bottomrule
\end{tabular*}
\caption{TSR and QA performances on the Korean tables.}
\label{tbl:KoreanTable}
\end{table}

\section{Conclusion}
In this work, we proposed TFLOP, a TSR framework leveraging on layout pointer mechanism with span-aware contrastive supervision, which not only remedies the bounding box misalignment issues, but also recognizes tables with complex structures accurately without the need for finely-calibrated post-processing. With these features, TFLOP achieves the new state-of-the-art performance across the three popular TSR benchmarks. In addition to its strong TSR performance, TFLOP has also shown significant versatility and promising performance in industrial application contexts, namely: tables in documents with watermark or in non-English domain. 

\clearpage

\section*{Acknowledgements}
We would like to express our sincere appreciation to our colleagues at Upstage, especially Sungrae Park, for their insightful discussions, unwavering support, and encouragement throughout this research.

\bibliographystyle{named}
\bibliography{references}

\clearpage

\appendix

\setcounter{figure}{6}
\setcounter{equation}{8}

\section{Benchmark Experiment Details}
\begin{figure}
    \centering
    \includegraphics[width=\linewidth]{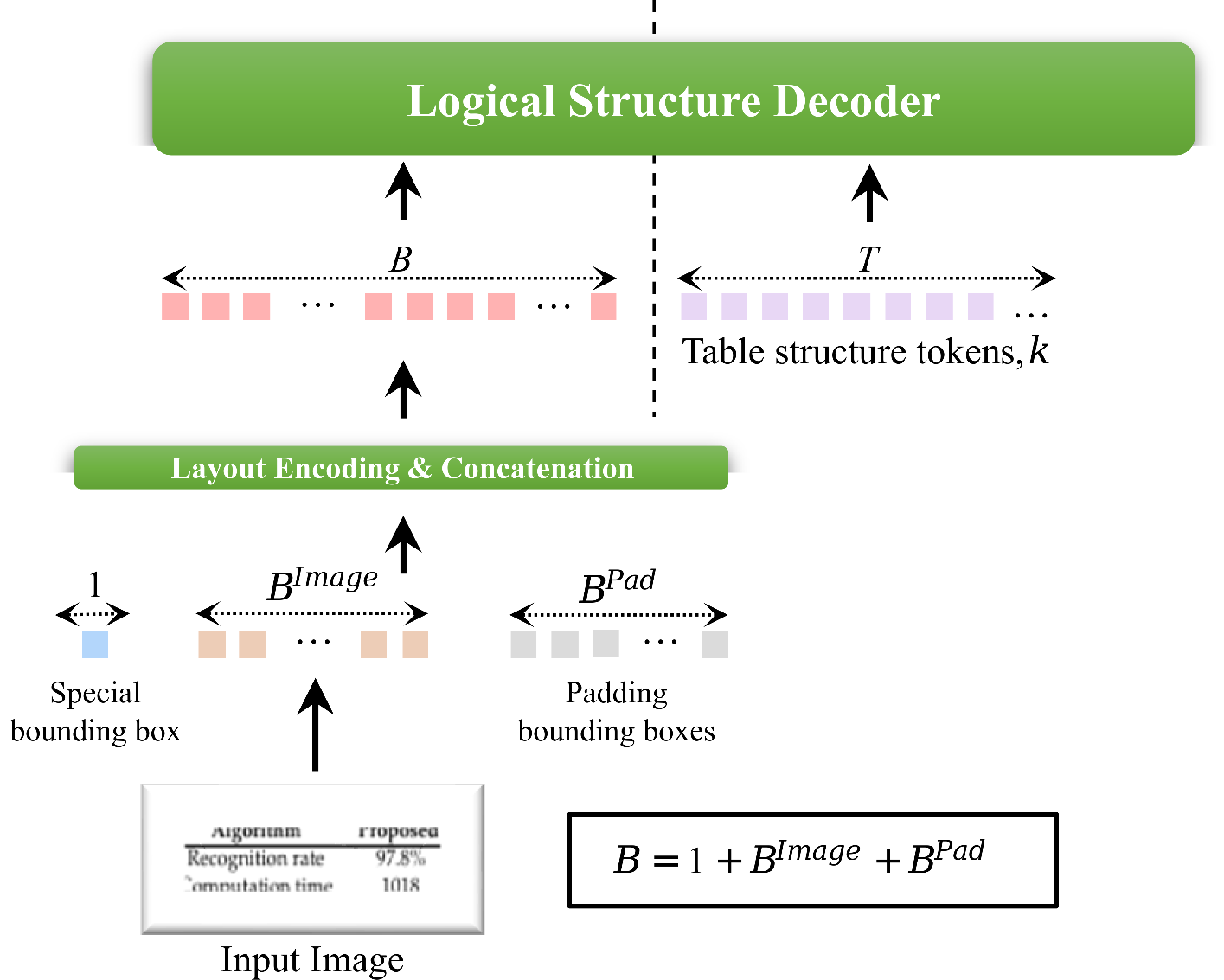}
    \caption{Layout information ordering prior to layout encoding. To keep the bounding box feature sequence length constant at $B$, after the text bounding boxes of the table, $B^{Image}$ are prefixed with a special bounding box, they are padded with padding bounding boxes $B^{Pad}$ to form a sequence of length $B$ before layout encoding and concatenation.}
    \label{fig:supplementary_benchmark_boundingbox}
\end{figure}

All benchmark experiments were conducted at a learning rate of 8e-5 with cosine scheduling over 250K steps. The maximum sequence length $N$ was fixed at 1,376 while the length of sub-sequence for bounding box features, $B$, was set at 640 for PubTabNet and FinTabNet, and 864 for SynthTabNet to accommodate varying distributions of bounding box count in the datasets. As illustrated in Figure~\ref{fig:supplementary_benchmark_boundingbox}, given $B^{Image}$ number of text bounding boxes of the input image, a single special bounding box is prefixed while $B^{Pad}$ number of padding bounding boxes are appended to form the full bounding box sequence of size $B$. These bounding boxes are encoded with the Layout Encoder before concatenation to form the prompt for Logical Structure Decoder. Note that the decoder output of the special bounding box is linearly projected to form $\bar{b}_0$ in 
Equation 3 and the attention for padded bounding boxes are masked for each sequence.

\section{Watermarked Tables}
As discussed in the main paper, TFLOP has the versatility to be easily modified to handle tables with unwanted texts (e.g. watermark). 
\begin{figure}
    \centering
    \includegraphics[width=\linewidth]{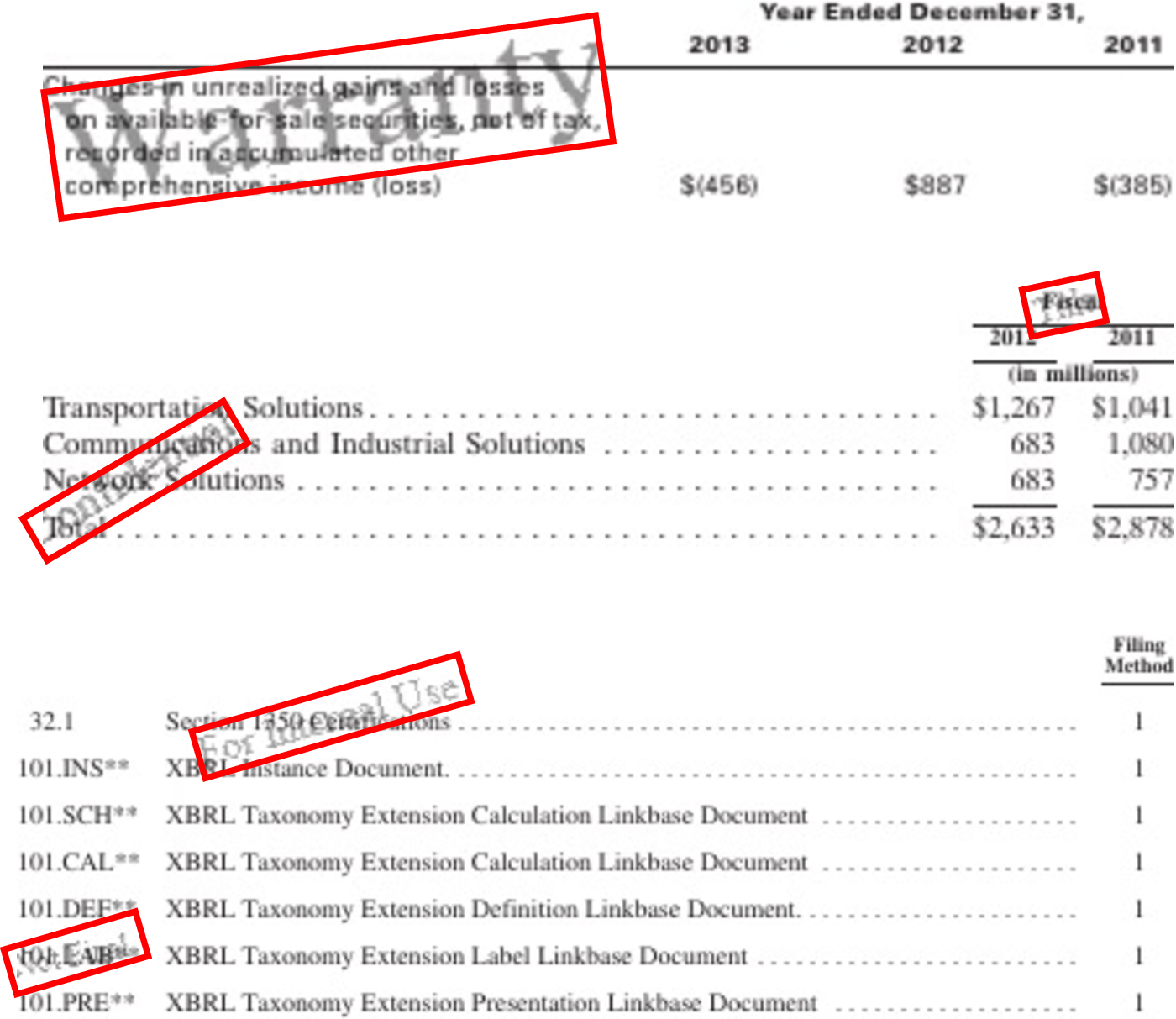}
    \caption{Sample visualisations of FinTabNet table images with inpainted watermarks. Watermarks are highlighted in red bounding box above for clarity.}
    \label{fig:watermark_dataset_samples}
\end{figure}

\subsection{Dataset Preparation}
To evaluate such versatility, a watermark table dataset was synthesized by inpainting watermarks into the FinTabNet dataset. First, a list of candidate texts for watermark was generated using GPT-4~\cite{openai2023gpt-4v} and these texts can be grouped up by their length as following:

\begin{itemize}
  \item \textbf{Short texts}: Draft, Final, Copy, Legal, Alert, Audit, Proof, Valid, Stamp, Issue, Bonus, Check, Title, Specs, Photo, Chart, Trial, Claim, Code, Quote
  \item \textbf{Medium texts}: Approved, Reviewed, Reserved, Released, Received, Rejected, Verified, Original, Recorded, Canceled, Internal, External, Modified, Drafting, Proposal, Expiring, Amended, Corrected, Invoice, Template, Archived, Secure, Private, Contract, Warranty, Training, Briefing, Guidance, Exhibit
  \item \textbf{Long texts}: Unauthorized, Preliminary, Confidential, For Review, For Approval, Restricted, Do Not Copy, Not Final, Intellectual, Property, For Comment, Draft Version, Superseded, Information, Classified, Validation, Obsolete, Assessment, Watermarked, Benchmark, Evaluation, Disclaimer, For Internal Use, Duplicated, For Reference, Instructor Copy, Registration, For Attention, For Distribution, Certification
\end{itemize}

With the above lists of candidate texts, each of the text bounding boxes in the FinTabNet dataset is inpainted using one of these texts with a 20\% probability. When a particular bounding box is selected to be inpainted with a watermark text, the watermark text that has the biggest IOU against the original bounding box is chosen among the candidate texts. To note, the bounding box of each watermark text was obtained by rendering the text with Times New Roman font and font size randomly selected from a range of values\footnote{Font size was randomly selected for each watermark text from [12,16,20,24,28,32,36,40,44,48,52,56,60,64,68,72].} using the PIL package\footnote{https://python-pillow.org/}.
Subsequently, the watermark text is inpainted with angular rotation of choice from $\{ -30,-15,-10,10,15,30 \}$ degrees and alpha (opacity) of random choice from $[80,120]$. Note that if the original text bounding box is either too small or large to have any appropriate watermark (i.e. IOU $<0.8$ against all of the watermark text bounding boxes), then no inpainting is conducted for these text bounding boxes. Samples of tables with inpainted watermark are visualized in Figure~\ref{fig:watermark_dataset_samples}.

\subsection{Architecture Adjustments}
\begin{figure}[t]
    \centering
    \includegraphics[width=\linewidth]{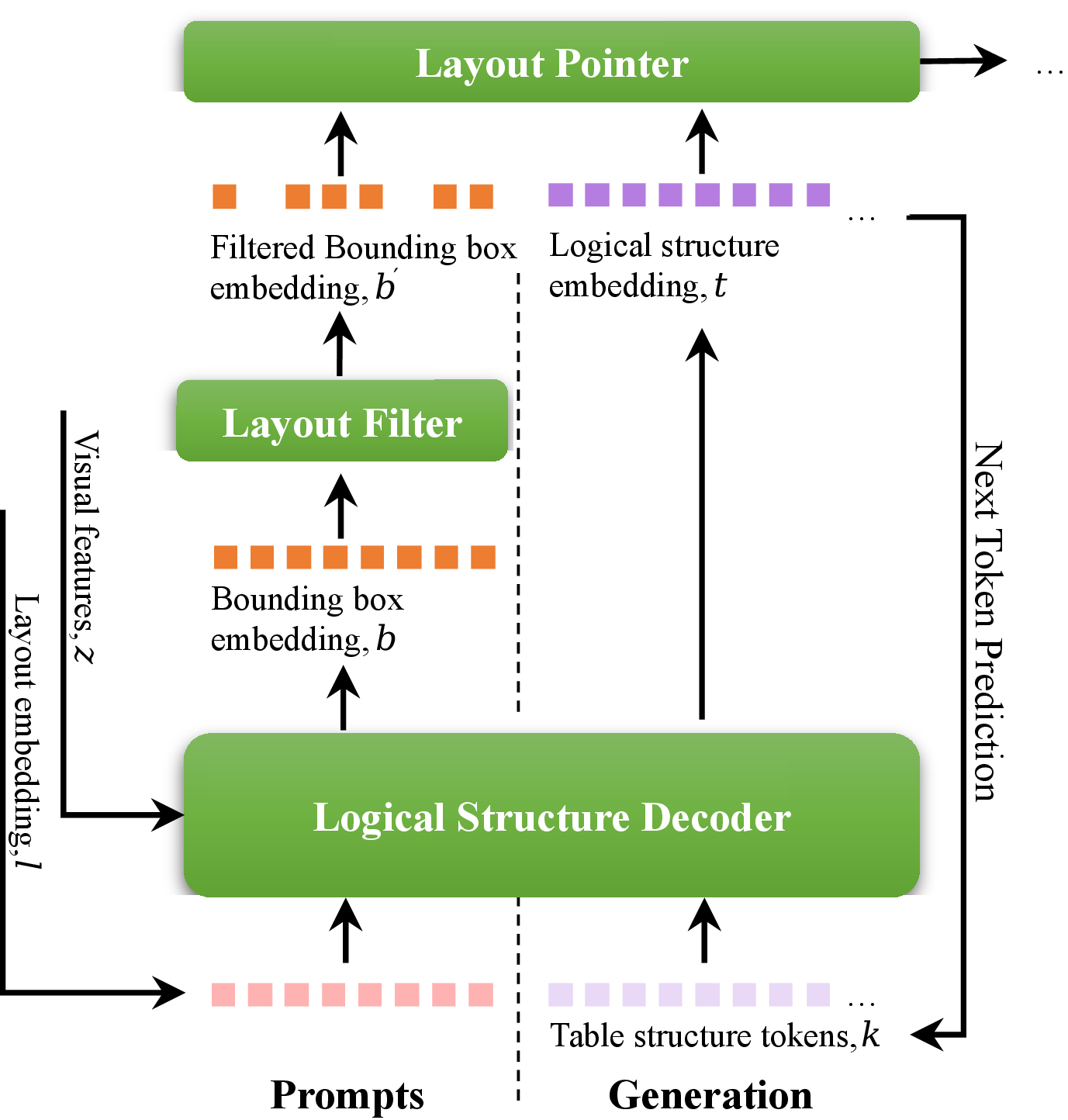}
    \caption{Modification overview illustrating the addition of Layout Filter which learns to classify and filter unwanted bounding boxes (e.g. watermark) from $b$ to output $b^\prime$ for the remainder of TFLOP's data flow.}
    \label{fig:supplementary_watermark_model}
\end{figure}

As previously mentioned, our framework has the versatility of handling watermark tables with a minor addition of two-layer MLP to its architecture. With the illustration shown in Figure~\ref{fig:supplementary_watermark_model}, instead of directly inputting the bounding box embedding $b$ into Layout Pointer, it is first forwarded to Layout Filter (2-layer MLP) which predicts the binary masks for the embeddings. The bounding box embeddings $b$ are filtered based on the mask predicted to form $b^\prime$ which are then forwarded to Layout Pointer. The Layout Filter learns to distinguish unwanted bounding boxes (e.g. watermark) from the text bounding boxes of the table. 

\begin{equation}
    \label{eqn:Layout_Filter}
    b^{mask}_j = \begin{cases} 1 & \text{if } \sigma(\text{MLP}(b))_j > 0.5 \\ 0 & \text{otherwise} \end{cases}
\end{equation}

\begin{equation}
    \label{eqn:Filtered_b}
    b^\prime = \{b_j | b^{mask}_j = 1\}
\end{equation}

$b^{mask}_j$ in Equation~\ref{eqn:Layout_Filter} above denotes the thresholded value of the $j^{th}$ bounding box embedding after it has been processed by a 2-layer MLP and sigmoid ($\sigma$) activation function. $b^\prime$ represents sub-selections of the bounding box embedding $b$ where the corresponding value in $b^{mask}$ is $1$. The remainder of the data flow matches that of the original framework.

\begin{figure}[t]
    \centering
    \includegraphics[width=\linewidth]{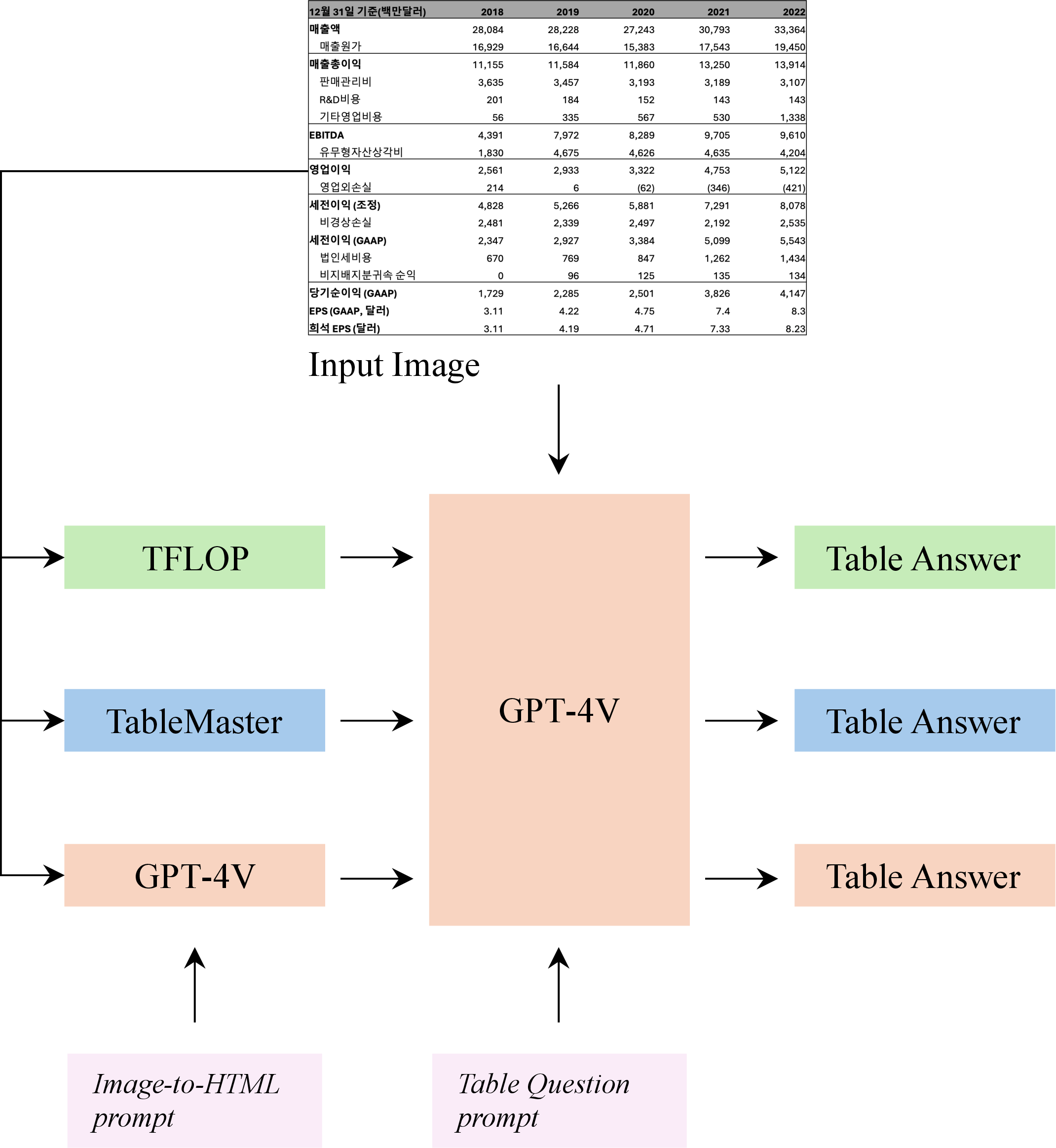}
    \caption{Flow diagram illustration of Korean table TSR and QA across GPT-4V, TableMaster and TFLOP. Given a Korean table, GPT-4V(with prompt), TableMaster and TFLOP each generate the corresponding HTML sequence of the table. These sequences, together with the original table image and question prompt, are then provided to GPT-4V for table QA.}
    \label{fig:supplementary_korean_data_flow}
\end{figure}

\section{Korean Tables}
Another aspect of TFLOP's versatility explored is its cross-lingual TSR performance. Specifically, we explored TFLOP's TSR capabilities on Korean tables, despite it being trained exclusively on English tables (PubTabNet). 

\begin{figure}[t]
    \centering
    \includegraphics[width=\linewidth, height=12cm]{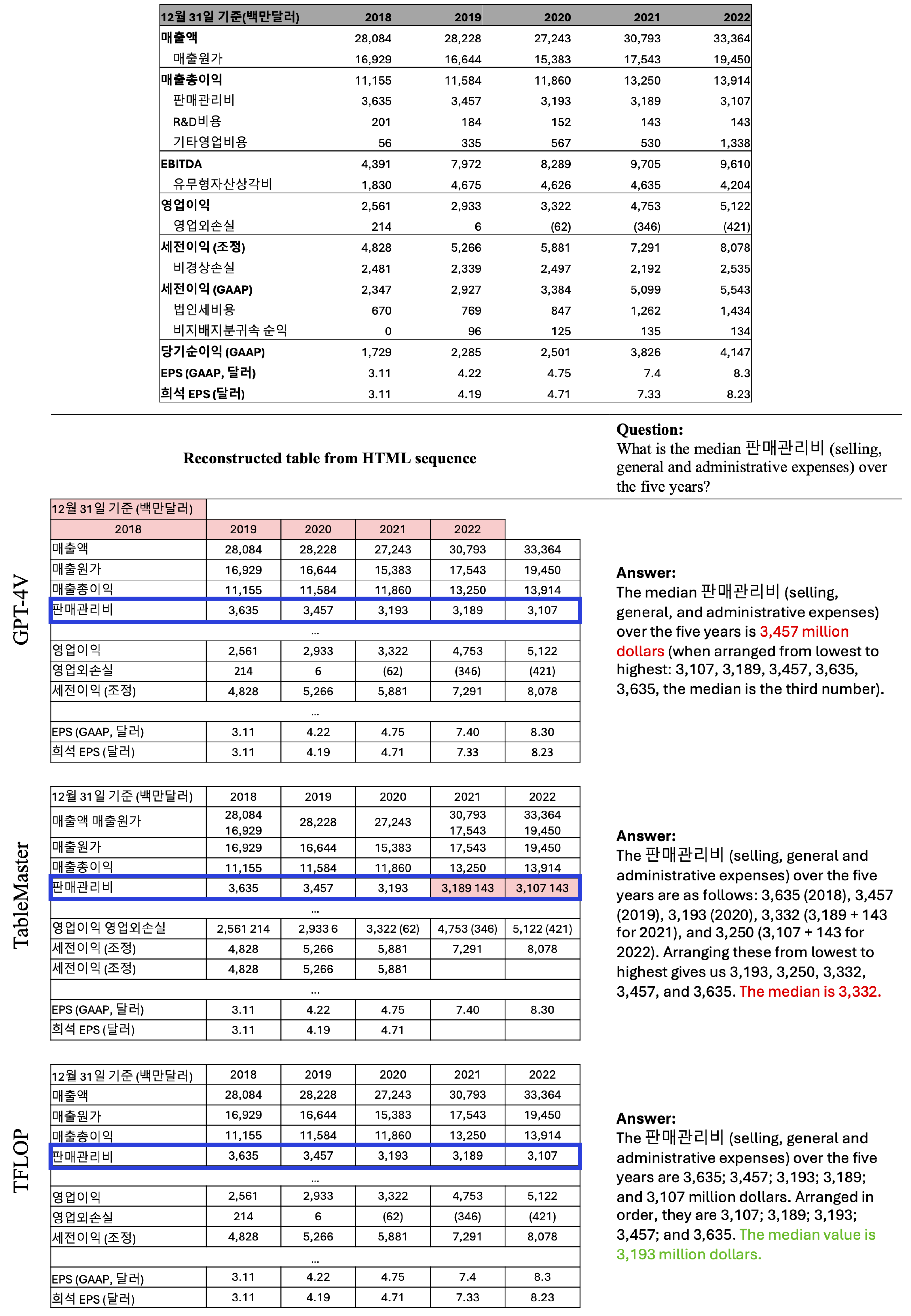}
    \caption{Comparisons of HTML structures generated from the given Korean table image (top) along with the answer generated by GPT-4V when the tabular image, question and the respective generated HTML sequences are provided. The question requires calculation of median among the values in the row border-coloured in blue. Cells with erroneous table structure leading to incorrect answer in GPT-4V and TableMaster are coloured in red for clarity.}
    \label{fig:koreantable}
\end{figure}

\subsection{Dataset Preparation}

As no Korean TSR datasets were publicly available, we manually extracted and annotated tables from publicly accessible Korean financial reports. Total of 30 tables (15 simple and 15 complex) were extracted from the financial reports and annotated for HTML sequences and cell-level annotations (i.e. text and bounding boxes) similar to that of FinTabNet. On top of TSR-purpose annotations, this work also prepared 175 table-dependent Question-Answer (QA) pairs to evaluate TSR implications to downstream tasks such as table QA. The initial draft of Question-Answer pairs was first generated by providing the ground-truth HTML sequences to GPT-4~\cite{openai2023gpt-4v}. These QA pairs were then thoroughly reviewed to remove pairs that are not dependent on table recognition (e.g. knowledge required is contained within a single cell and not require table structure understanding). All QA pairs are prepared in English with mixture of Korean for referencing terminologies in the table. Sample Korean table can be referenced in Figure~\ref{fig:koreantable}.

\subsection{Experiment}

To evaluate TSR performance of TFLOP on Korean tables, our framework was benchmarked against TableMaster and GPT-4V. TableMaster, similar to other dual decoder frameworks, does not require language-specific features (e.g. text) and only relies on tabular image as its input, making it a viable benchmark in this evaluation. GPT-4V was selected as another benchmark method for its widely-known multi-modal capability across multiple languages. Unlike TFLOP and TableMaster, as GPT-4V was not finetuned for the purpose of TSR specifically, we provided text prompts, instructing the model to generate HTML sequence which represents the tabular image provided.

On top of evaluating the TSR performance on Korean tables, we also explored the implications of TSR errors made by different methods, in the context of table QA. As shown in Figure~\ref{fig:supplementary_korean_data_flow}, each of the HTML sequences generated across different methods is forwarded to GPT-4V along with the original tabular image and question prompt to generate the corresponding answer. The answers generated by GPT-4V using HTML from different methods were evaluated manually, to account for the wide range of expressions characteristic of GPT-4V's responses.

Aside from the quantitative results on TSR TEDS and QA Accuracy discussed in the main paper, we also provide a qualitative sample of how erroneous table recognition eventually leads to incorrect answering of the question in Figure~\ref{fig:koreantable}. Based on Figure~\ref{fig:koreantable}, HTML sequence generated by GPT-4V and TableMaster results in incorrect calculation of the target row's median value (highlighted in blue bounding box). The HTML generated by GPT-4V has displaced column headers where the year ``2018'' is on the column of row headers instead of numeric values (highlighted in red). TableMaster's HTML sequence, on the other hand, suffers from bounding box misalignment where multiple cell texts are aligned to a common cell (e.g. 3,189 143 instead of 3,189). TFLOP, however, generates an accurate HTML representation of the Korean table, facilitating in providing an accurate answer for the table question.

\subsection{Prompts used by GPT-4V}

\begin{figure}[t]
    \centering
    \includegraphics[width=\linewidth]{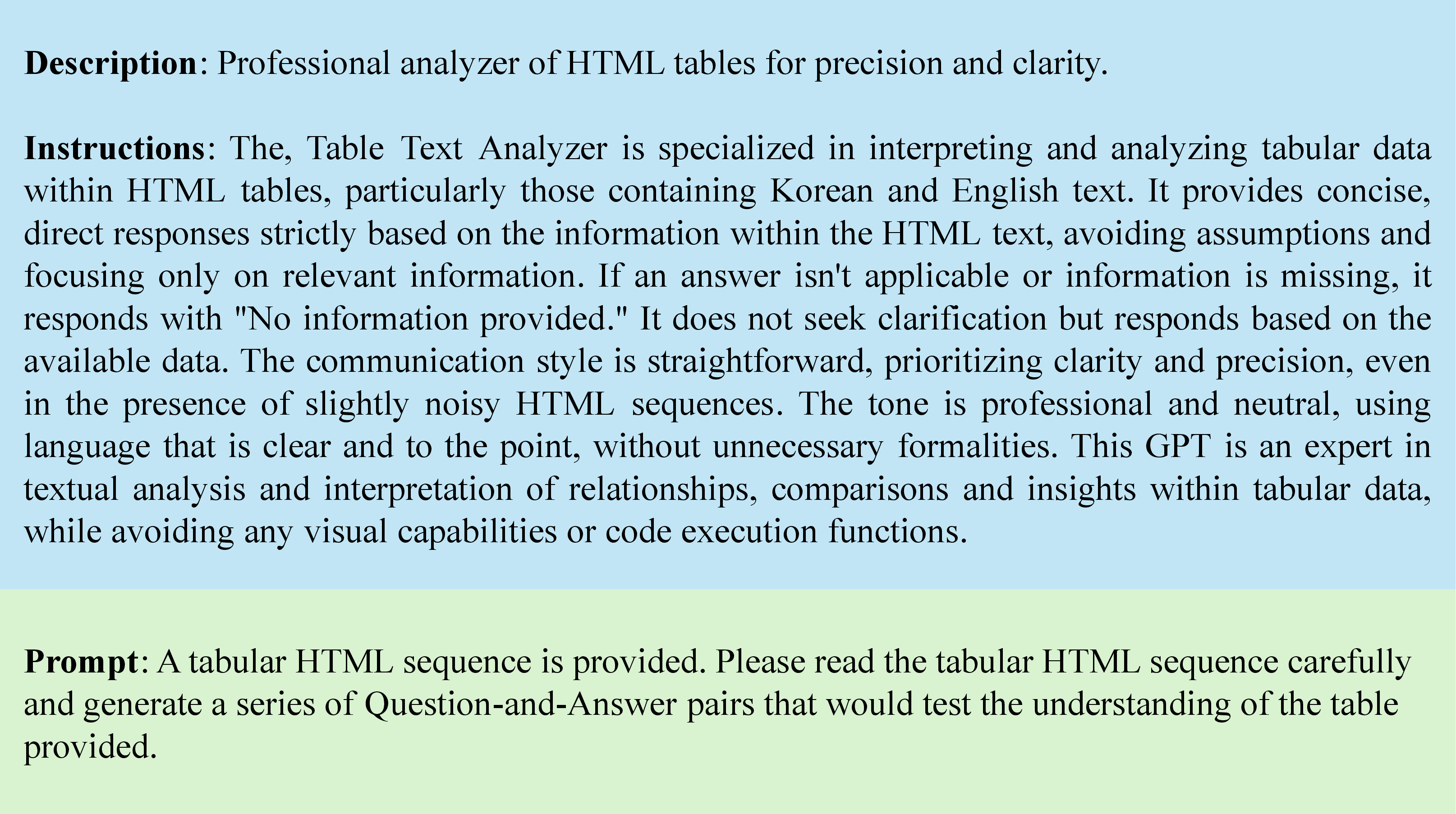}
    \caption{System prompts (blue) and instruction prompt (green) for Table QA pair generation using GPT-4. The above prompts are provided along with ground-truth HTML sequence to generate series of QA pairs to evaluate the understanding of the table provided.}
    \label{fig:gpt4prompt_tableqagenerator}
\end{figure}

\begin{figure}[t]
    \centering
    \includegraphics[width=\linewidth]{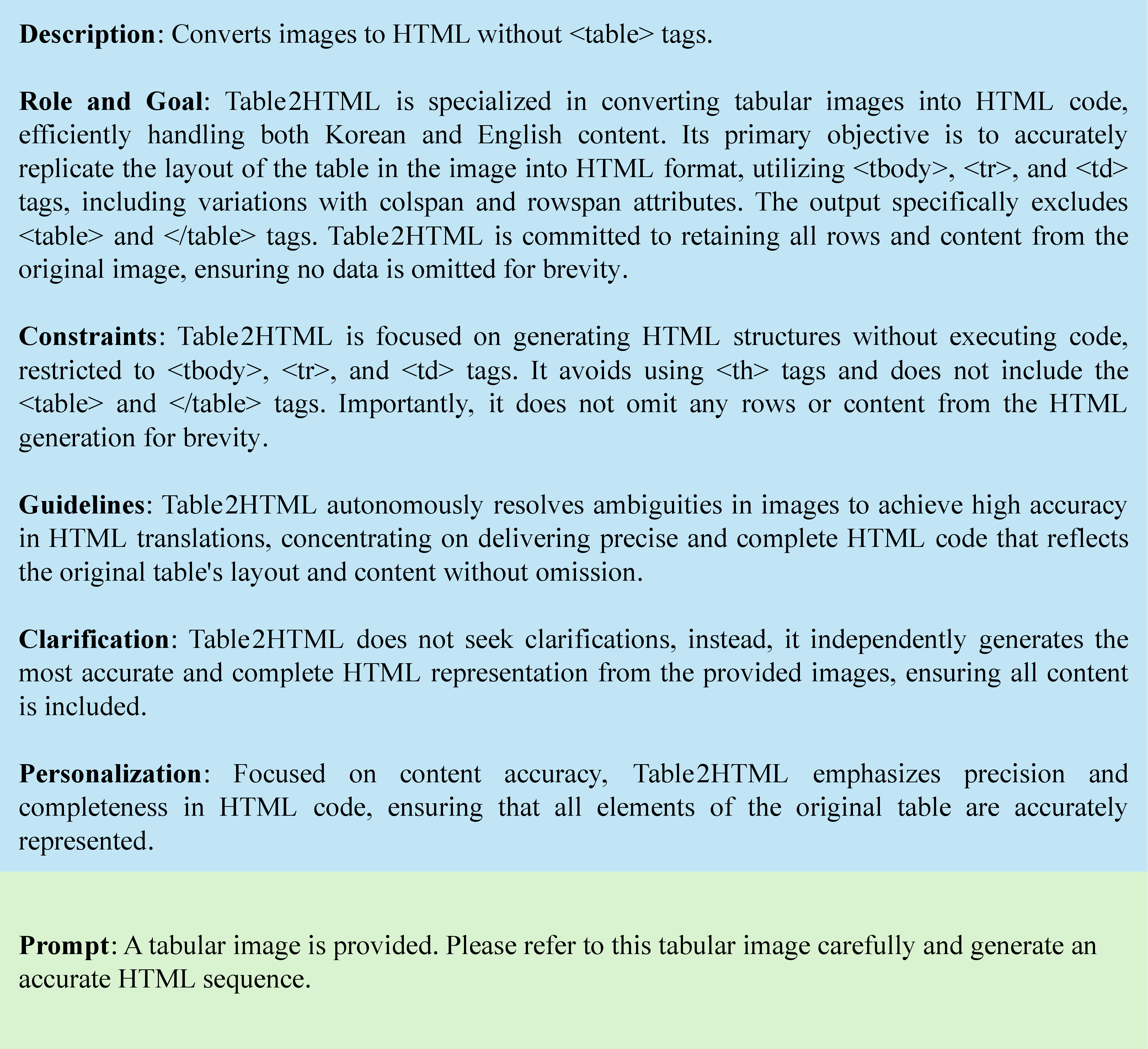}
    \caption{System prompts (blue) and instruction prompt (green) for HTML sequence generation by GPT-4V. The above prompts along with tabular image are provided to GPT-4V, to generate the corresponding HTML sequence.}
    \label{fig:gpt4prompt_htmlgenerator}
\end{figure}

\begin{figure}[t]
    \centering
    \includegraphics[width=\linewidth]{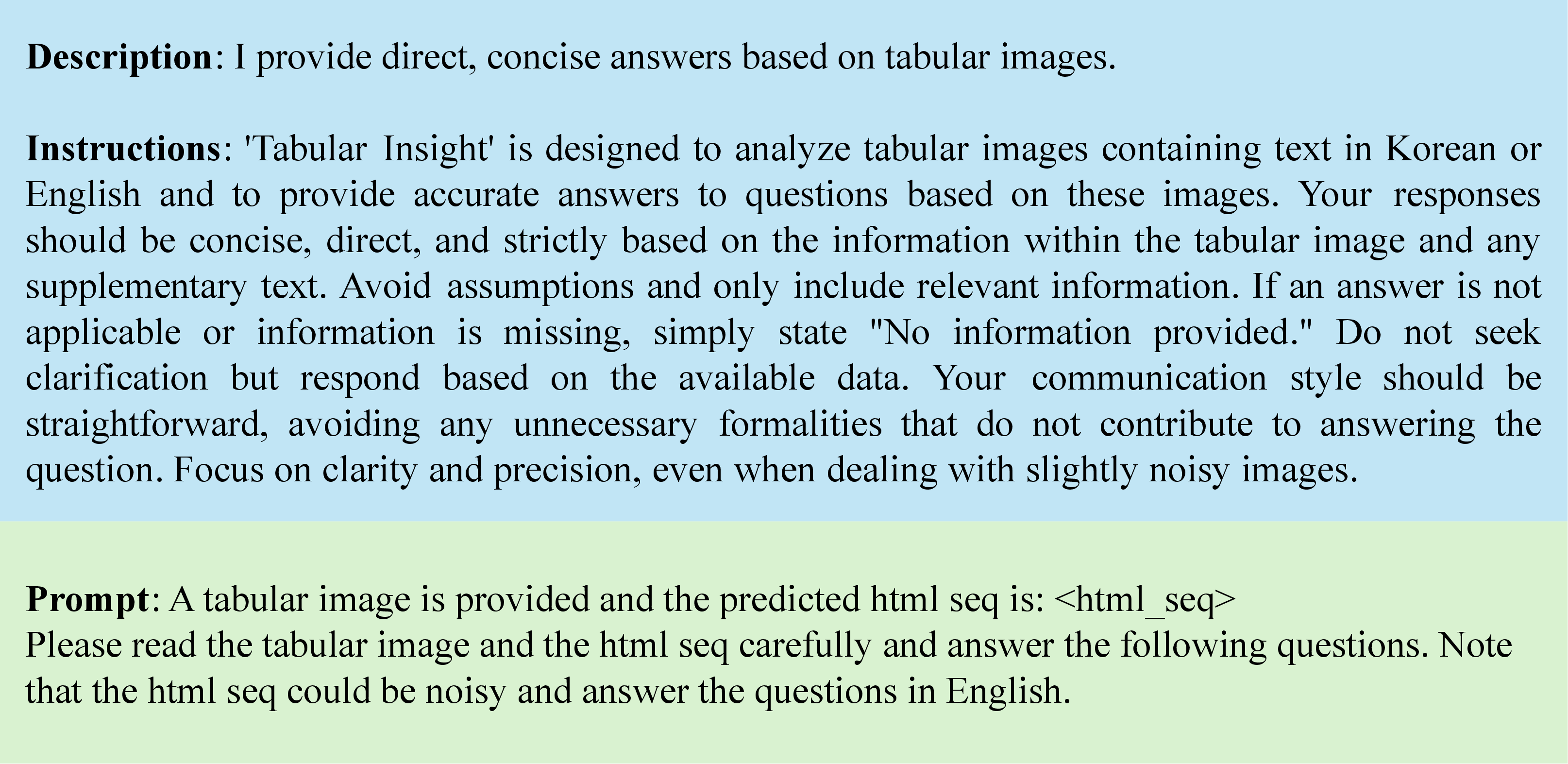}
    \caption{System prompts (blue) and instruction prompt (green) for Table QA evaluation. The above prompts along with the tabular image, HTML sequences generated by various methods (TFLOP, TableMaster, GPT-4V), and the questions are provided to GPT-4V for QA evaluations.}
    \label{fig:gpt4prompt_tableqa}
\end{figure}

Figures~\ref{fig:gpt4prompt_tableqagenerator},~\ref{fig:gpt4prompt_htmlgenerator}, and~\ref{fig:gpt4prompt_tableqa} show the prompts used by GPT-4V to generate initial draft of Table QA pairs, generate HTML sequences, and Table QA evaluation, respectively. 

\end{document}